\documentclass{article}

\usepackage{microtype}
\usepackage{graphicx}
\usepackage{booktabs} %

\usepackage{hyperref}

\usepackage[accepted]{preprint_custom}

\usepackage{amsmath}
\usepackage{amssymb}
\usepackage{mathtools}
\usepackage{amsthm}

\usepackage[capitalize,noabbrev]{cleveref}

\theoremstyle{plain}

\theoremstyle{definition}

\theoremstyle{remark}

\usepackage[textsize=tiny]{todonotes}

\graphicspath{{figures/}}

\preprinttitlerunning{Block-Toeplitz Covariance Matrices for LDA in ERP-based BCI}

\usepackage[utf8]{inputenc} %
\usepackage[T1]{fontenc}    %
\usepackage{hyperref}       %
\usepackage{url}            %
\usepackage{booktabs}       %
\usepackage{nicefrac}       %
\usepackage{microtype}      %
\usepackage{marginnote}

\expandafter\let\csname equation*\endcsname\relax

\expandafter\let\csname endequation*\endcsname\relax

\usepackage{mathtools}
\usepackage{amssymb}
\usepackage{amsfonts}       %
\usepackage{blkarray}
\usepackage{siunitx}
\usepackage{subcaption}

\usepackage{xstring}
\usepackage{catchfile}

\usepackage[printwatermark]{xwatermark}
\usepackage{xcolor}
\usepackage{graphicx}
\graphicspath{{figures/}}

\usepackage{multirow}

\usepackage[textsize=tiny]{todonotes}

\usepackage{url}

\reversemarginpar

\renewcommand{\vec}[1]{\boldsymbol{\mathbf{#1}}}

\DeclareMathOperator{\cov}{cov}

\newcommand{\trans}{^\mathsf{T}}
\newcommand{\mat}[1]{\boldsymbol{\mathbf{#1}}}

\newcommand{\datTr}{\mathcal{D}_\text{tr}}
\newcommand{\datVal}{\mathcal{D}_\text{val}}

\DeclareMathSymbol{\shortminus}{\mathbin}{AMSa}{"39}
 
\begin{document}

\widowpenalty 10000
\clubpenalty 10000
\twocolumn[
\preprinttitle{Introducing Block-Toeplitz Covariance Matrices to Remaster Linear Discriminant Analysis for Event-related Potential Brain-computer Interfaces}
\preprintsetsymbol{equal}{*}

\begin{preprintauthorlist}
\preprintauthor{\textbf{Jan Sosulski} \small{Department of Computer Science, University of Freiburg, Freiburg, Germany}}{uni_freiburg}
\preprintauthor{\textbf{Michael Tangermann} \small{Donders Institute for Brain, Cognition and Behaviour, Radboud University, Nijmegen, The Netherlands}}{uni_netherlands}
\preprintauthor{\emph{Correspondence to:} \texttt{michael.tangermann@donders.ru.nl}}{asd}
\end{preprintauthorlist}

\preprintkeywords{Brain signal classification, linear discriminant analysis, high dimensional covariance estimation, block-Toeplitz matrix, spatiotemporal data}

\vskip 0.3in
]

\begin{abstract}
\textbf{\emph{Abstract}---}Covariance matrices of noisy multichannel electroencephalogram time series data are hard to estimate due to high dimensionality.
In brain-computer interfaces (BCI) based on event-related potentials and a linear discriminant analysis (LDA) for classification, the state of the art to address this problem is by shrinkage regularization.
We propose a novel idea to tackle this problem by enforcing a block-Toeplitz structure for the covariance matrix of the LDA, which implements an assumption of signal stationarity in short time windows for each channel.
On data of 213 subjects collected under 13 event-related potential BCI protocols, the resulting `ToeplitzLDA' significantly increases the binary classification performance compared to shrinkage regularized LDA (up to 6 AUC points) and Riemannian classification approaches (up to 2 AUC points).
This translates to greatly improved application level performances, as exemplified on data recorded during an unsupervised visual speller application, where spelling errors could be reduced by 81\,\% on average for 25 subjects.
Aside from lower memory and time complexity for LDA training, ToeplitzLDA proved to be almost invariant even to a twenty-fold time dimensionality enlargement, which reduces the need of expert knowledge regarding feature extraction.

 \end{abstract}

\section{Introduction}
\label{sec:introduction}

A brain-computer interface (BCI) enables the use of brain activity as an input for a computer program.
Example applications are spelling programs~\cite{farwell1988talking}, wheelchair control~\cite{fernandez-rodrguez2016review} or even rehabilitation trainings for neurological diseases, e.g., stroke~\cite{ang2013brain-computer} or attention deficit hyperactivity disorder~\cite{lim2012brain-computer}.
As brain activity differs for each individual, modern BCIs make use of machine learning methods to adapt to each user.
To record brain activity, the electroencephalogram (EEG) is a popular choice, as it is non-invasive, inexpensive and almost anyone can use it~\cite{nunez2012electric}.
The decoding of EEG signals using machine learning methods is challenged by a very low signal-to-noise ratio of the EEG and as the characteristics of the EEG signal ideally should be learned from very little individual training data.
One approach to solve this problem is to use transfer learning~\cite{jayaram2016transfer}.
However, this can be hard to realize as the EEG shows large differences between subjects and even between sessions of the same subject.
Another approach is to make classifiers more data efficient, i.e., by exploiting better the available data~\cite{,gruenwald2019time-variant,sosulski2021improving}.

One commonly used type of brain activity for BCIs is the event-related potential (ERP), a transient potential as a response to a specific stimulus, e.g., a visual emphasis of letters on a screen~\cite{farwell1988talking,lin2018novel} or playing a tone or word~\cite{schreuder2010new,hohne2012natural}.
Here the BCI user is presented with multiple stimuli while the BCI exploits the fact that attended \emph{target} stimuli produce different ERPs than ignored \emph{non-target} ones~\cite{sellers2012bcis}.
By classifying the ERP responses, the target stimulus, e.g., the attended letter on a screen, can be detected.
Note that the amplitude of an ERP is often more than ten times smaller than that of the ongoing background activity in the EEG.
For classification of ERP components, linear discriminant analysis (LDA) is often used, as it is computationally cheap during training and inference and still delivers state of the art performance~\cite{lotte2007review,lotte2018review}.
LDA requires estimates of target and non-target class means and of the feature distribution expressed by covariances.
As BCI problems often come with few training data points, the difficult estimation of the potentially high-dimensional covariance matrix is typically regularized by shrinking~\cite{ledoit2004well-conditioned,zhao2017oracle,blankertz2011single-trial}.

ERPs are analyzed by extracting epochs, i.e., short time windows aligned to the onset of each stimulus.
While EEG generally exhibits non-stationarity over longer time scales, a weak stationarity assumption, i.e., that mean and autocovariance do not depend on time, is reasonable~\cite{cohen1977stationarity} for short (few seconds) epochs of high-pass filtered EEG.
While an ERP response can be seen as a mean shift non-stationarity, it is removed with other mean information prior to estimating the covariance matrix.
For data generated from a stationary process, the estimation of the covariance matrix becomes much easier~\cite{xiao2012covariance}, especially in high-dimensional settings~\cite{chen2013covariance}.
Another popular method to deal with time series covariance estimation is tapering~\cite{furrer2007estimation,pourahmadi2013banding} where the entries of the covariance matrix are regularized towards zero with increasing temporal distance.

We propose to make LDA training for spatiotemporal ERP features more efficient by enforcing these assumptions.
We hypothesize---and later show in ERP data and classification results---that imposing a block-Toeplitz structure on the covariance matrix of the background activity of the EEG in ERP epochs can greatly enhance the classification performance of LDA.
\section{Methods}
\label{sec:methods}

\subsection{Benchmark Datasets}
\label{sec:datasets}

We use datasets with ERPs evoked by stimuli modalities based on either words, tones, or visual stimuli.
The signal-to-noise ratio (the ERP amplitude) depends, among others, on this stimulus modality.
In the context of LDA, class means are provided by the average target/non-target ERPs, whereas background activity and other noise are characterized by the covariance matrix.

\begin{table}
	\centering
	\caption{ERP datasets used for the benchmark. Note that datasets marked with \textsuperscript{p} are private and datasets with \textsuperscript{*} have less than 384 epochs available in the training set (see~\Cref{sec:benchmark_bin}).}
	\label{tab:datasets}
	\begin{tabular}{lllll}
		\toprule
		\textbf{Dataset}                     & \textbf{Modality} & \textbf{Subjects} & \textbf{Channels} \\
		\midrule
		V\_BI                                & visual            & 24                & 16                \\
		V\_BNCI\_A                           & visual            & 8                 & 8                 \\
		V\_BNCI\_B                           & visual            & 8                 & 8                 \\
		V\_BNCI\_C\textsuperscript{*}        & visual            & 10                & 16                \\
		V\_EPFL                              & visual            & 7                 & 32                \\
		V\_Lee                               & visual            & 54                & 62                \\
		V\_LLP\_A                            & visual            & 13                & 31                \\
		V\_LLP\_B                            & visual            & 12                & 31                \\
		T\_Odd\_Healthy\textsuperscript{p,*} & tone              & 20                & 63                \\
		T\_Odd\_Stroke\textsuperscript{p,*}  & tone              & 14                & 31                \\
		T\_Odd\_Young                        & tone              & 13                & 31                \\
		W\_Healthy\textsuperscript{p}        & word              & 20                & 63                \\
		W\_Stroke\textsuperscript{p}         & word              & 10                & 31                \\
		\bottomrule
	\end{tabular}
\end{table}

For a general overview of all used datasets please refer to~\Cref{tab:datasets}, while dataset details are provided in~\Cref{sec:supp:datasets}.
We used MOABB (Mother of all BCI Benchmarks)~\cite{jayaram2018moabb} to access publicly available datasets, enriched by private datasets recorded in our lab.

\subsection{Classification Pipeline}
\label{sec:classification_pipeline}

While multiple sessions were recorded for some subjects, we are mainly interested in within-session performance, i.e., training and applying a classifier to data of the same session.
Data of a session is split exclusively at boundaries of continuous runs, into a training set $\datTr$ and a validation set $\datVal$, with $|\datTr| \approx |\datVal|$.
The continuous runs were band-pass filtered to a typical ERP frequency range of $[0.5, 16]$\,Hz.
We then obtained feature matrices $\mat{X}$ by windowing epochs relative to the onset of all stimuli and resampling the epochs to $40$\,Hz.
A typical ERP pipeline predefines time intervals in which the signal is averaged.
Thus we evaluate different ERP intervals and report the one performing best across all datasets.
In addition, we evaluate a setting where every single time sample in a specific time window is used.
For the complete list of used hyperparameters see~\Cref{sec:supp:hyperpars}.

\subsection{Classification with Linear Discriminant Analysis}

In a binary classification task, the Fisher linear discriminant analysis (LDA) requires two statistical measures of the data~\cite{bishop2006linear}:
First, the within-class covariance matrix $\mat{\Sigma}$ and second the class means $\vec{\mu_1}$ and $\vec{\mu_2}$.
With this, we can calculate the weight vector $\vec{w} = \mat{\Sigma}^{\shortminus 1}(\vec{\mu}_2-\vec{\mu}_1)$ and the bias $b = \vec{w}\trans (\vec{\mu}_1+\vec{\mu}_2)$.
For high-dimensional features, shrinkage regularized covariance matrices~\cite{ledoit2004well-conditioned,blankertz2011single-trial} are used in practice to improve conditioning.

While generally a bias adaptation is required for unbalanced classes this is unnecessary for most ERP-based BCI applications, as the stimulus set contains (besides multiple non-targets) exactly one target stimulus.
Assuming positive target labels, it can safely be determined by the highest classifier output.
This may explain the popularity of the area under the receiver operating characteristic curve (AUC) as a metric in the realm of BCI, as the bias does not impact the AUC metric.

\subsection{Enforcing Block-Toeplitz Covariance Matrices for Stationary Multichannel Time Series Data}

\subsubsection{Preliminary: EEG Signal Generation}

During a typical ERP BCI experiment you usually record a continuous EEG signal at $N_c$ channels, which is then cut into $N_e$ epochs with $N_t$ time samples each.

The EEG also records signals unrelated to brain activity, e.g., amplifier noise or line frequency noise, and physiological artifacts caused by muscle activity.
While these artifacts can be suppressed in offline analysis, e.g., by discarding artifactual epochs, this may not be as straightforward in online settings.
Therefore, we decided against using artifact rejection and left artifacts in the data as additional challenges for the classifiers.

\subsubsection{Feature Extraction}

ERP features come from the spatial domain, i.e., the distribution of the EEG electrodes across the scalp, and the temporal domain, as the ERP is recorded over a time period.
This results in a feature matrix %
per epoch, which is usually flattened into a vector $\vec{x} \in \mathbb{R}^{N_c N_t}$.
Depending on which dimension represents row or column, we obtain
\begin{align*}
	\vec{x}   & \coloneqq [x_{c_1}^{t_1}, x_{c_2}^{t_1}, \ldots, x_{c_{N_c}}^{t_1}, x_{c_{1}}^{t_2}, \ldots, x_{c_{N_c}}^{t_{N_t}}] \hspace{5pt} \textrm{or} \\
	_t\vec{x} & \coloneqq [x_{c_1}^{t_1}, x_{c_1}^{t_2}, \ldots, x_{c_1}^{t_{N_t}}, x_{c_2}^{t_{1}}, \ldots, x_{c_{N_c}}^{t_{N_t}}],
\end{align*}

where $x_{c_m}^{t_n}$ is the signal at the $m$-th channel and the $n$-th time point of that epoch.

In this paper we use the form $\vec{x}$, where first all channel features are stacked, and refer to it as being \emph{channel-prime}, whereas we refer to $_t\vec{x}$ as \emph{time-prime}.
This choice impacts on the structure of the covariance matrix $\mat{\Sigma}$:
Stacking each feature vector of all $N_e$ epochs, we finally obtain our data $\mat{X}\in\mathbb{R}^{N_c N_t \times N_e}$.
Using the centered data $\mat{\tilde{X}}$, we can estimate the sample covariance matrix as $\frac{1}{N_e - 1}\tilde{\mat{X}}\tilde{\mat{X}}\trans$.

When using feature vectors $\vec{x}$ that are channel-prime, we can construct the channel-prime covariance matrix $\mat{\Sigma}$ as a block matrix consisting of multiple blocks of the form
\begin{align*}
	\mat{C}_{i, j} \coloneqq \begin{bmatrix}
		\cov({x}_{c_1}^{t_i},{x}_{c_1}^{t_j})       & \hdots & \cov({x}_{c_1}^{t_i},{x}_{c_{N_c}}^{t_j})     \\
		\vdots                                      & \ddots & \vdots                                        \\
		\cov({x}_{c_{N_c}}^{t_i},{x}_{c_{1}}^{t_j}) & \hdots & \cov({x}_{c_{N_c}}^{t_i},{x}_{c_{N_c}}^{t_j}) \\
	\end{bmatrix},
\end{align*}

where block $\mat{C}_{i, j} \in \mathbb{R}^{N_c \times N_c}$ describes the channel-wise (cross\nobreakdash-)\hspace{0pt}covariance matrix between the $i$-th and $j$-th time point.
Note that in this work we do not enforce any assumptions about the spatial covariance between channels and simply use the empirical values.
With this, we can write the complete channel-prime covariance matrix as
\begin{align*}
	\mat{\Sigma} \coloneqq & \begin{bmatrix}
		\mat{C}_{1,1}    & \mat{C}_{1,2} & \hdots & \mat{C}_{1,N_t}    \\
		\mat{C}_{2,1}    & \mat{C}_{2,2} &        & \vdots             \\
		\vdots           &               & \ddots & \vdots             \\
		\mat{C}_{N_t, 1} & \hdots        & \hdots & \mat{C}_{N_t, N_t}
	\end{bmatrix}                                        \\
	=                      & \begin{bmatrix}\mat{C}_{i,j}\end{bmatrix}_{1 \leq i \leq N_t, 1 \leq j \leq N_t} \\
\end{align*}
which describes the full relationship of the features across both the temporal and spatial domain.

\subsubsection{Enforcing Stationarity}
\label{sec:assumptions}

We now want to incorporate two assumptions for ERP data:

\begin{enumerate}
	\item[\textbf{A1}] Only the ERP signal is time locked, whereas the EEG background activity is generated from a stationary process~\cite{xiao2012covariance}. As a result, the covariance across the time domain depends only on the temporal distance of two samples, i.e., $\cov({x}^{t_j}, {x}^{t_i}) = \cov({x}^{t_j + \delta}, {x}^{t_i + \delta}) \; \forall \delta \in \mathbb{R}$. %
	\item[\textbf{A2}] For increasing temporal distances, i. e., $|t_i-t_j| \to \infty$ the covariance goes towards zero~\cite{pourahmadi2013banding}.
\end{enumerate}

Both assumptions only concern the temporal, but not the spatial domain. To incorporate assumption \textbf{A1} %
of stationarity into the channel-prime covariance matrix, i.e., $i-j = k-l \implies \mat{C}_{i,j} = \mat{C}_{k,l}$, we introduce the notation $\mat{C}_{\Delta t = d}$, where $d$ is the signed temporal distance. Now we can write the stationary covariance matrix as
\begin{align*}
	\mat{\Sigma}^{\text{stat}} \stackrel{\text{\textbf{A1}}}{\coloneqq} \begin{bmatrix}
		\mat{C}_{\Delta t = 0}                   & \mat{C}_{\Delta t = 1}                   & \hdots & \mat{C}_{\Delta t = N_t - 1} \\
		\mat{C}_{\Delta t = \shortminus 1}       & \mat{C}_{\Delta t = 0}                   & \hdots & \mat{C}_{\Delta t = N_t - 2} \\
		\vdots                                   & \vdots                                   & \ddots & \vdots                       \\
		\mat{C}_{\Delta t = \shortminus N_t + 1} & \mat{C}_{\Delta t = \shortminus N_t + 2} & \hdots & \mat{C}_{\Delta t = 0}       \\
	\end{bmatrix},
\end{align*}
which has a block-Toeplitz structure. Note that it is fully defined by its first block-row or block-column as the block-diagonals of $\mat{\Sigma}_{\text{stat}}$ are equal and as $\mat{C}_{\Delta t = \shortminus d}\trans = \mat{C}^{\phantom{T}}_{\Delta t = d}$ holds due to symmetry.

\begin{figure}[t]
	\centering
	\includegraphics[width=\columnwidth]{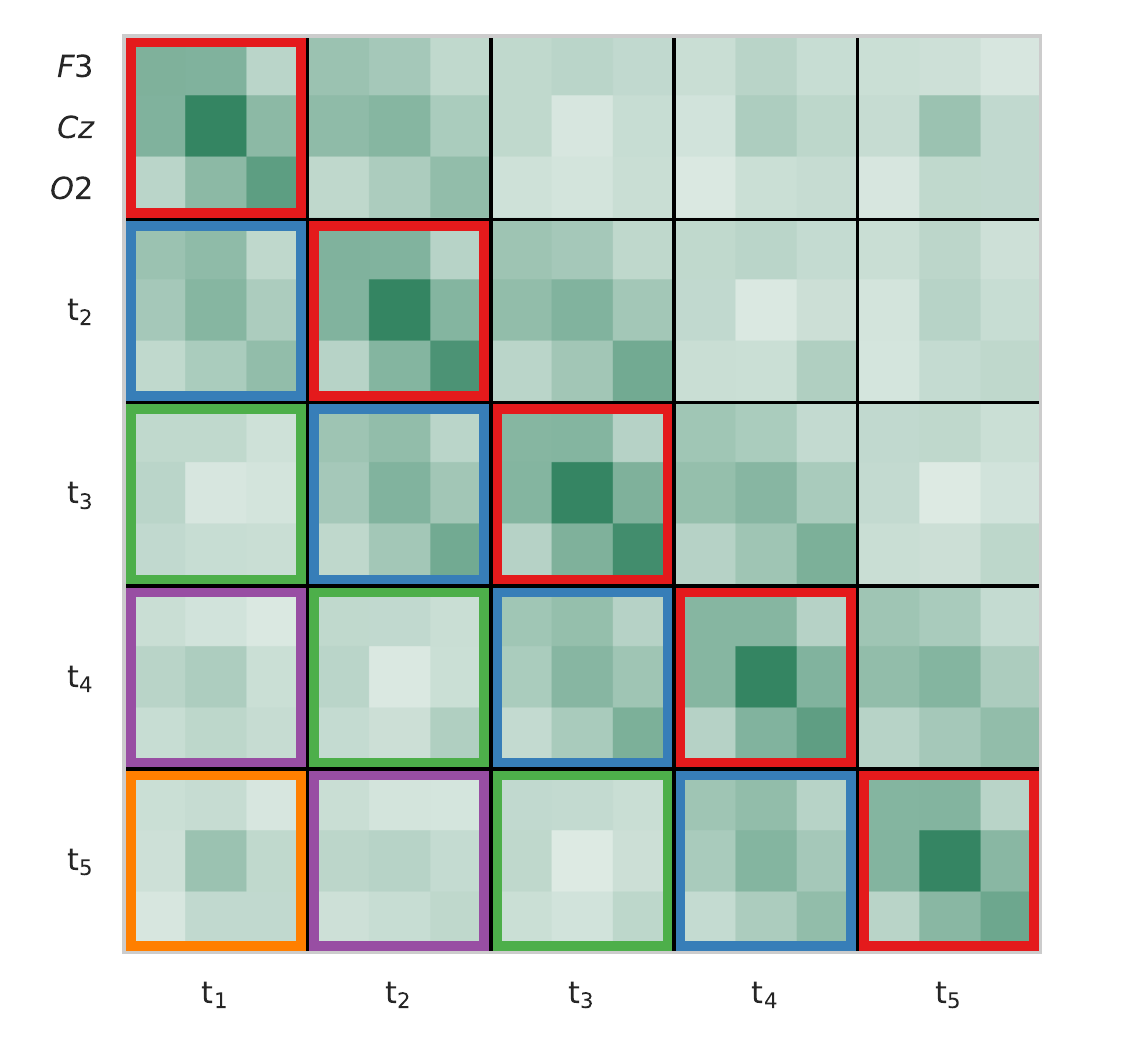}
	\caption{Example of a sample covariance matrix of an ERP feature vector for the time points [0.1, 0.15, 0.2, 0.25, 0.3]\,s with three EEG channels from different head locations: F3 (front left), Cz (center) and O2 (back right). The matrix was calculated on 3896 epochs of a visual ERP paradigm. Black lines delineate the channel-wise (cross-)covariance blocks of the matrix. In order to transform this general covariance matrix into block-Toeplitz form, you would average along the block-diagonals, i.e., averaging blocks sharing the same color.}
	\label{fig:cov_free_params}
\end{figure}

The advantage of representing the covariance matrix in this way is that the number of free parameters that need to be estimated decreases from $(N_t N_c)(N_t N_c+1)/2$, which scales quadratically in $N_t$, to only $N_c(N_c + 1)/2 + (N_t-1)N_cN_c$, which scales linearly in $N_t$.
Aside from fewer free parameters, using block-Toeplitz covariance matrices reduces memory usage, even if many temporal features are used.

For block-Toeplitz matrices computationally efficient algorithms for inversion or solving linear systems exist.
We sped up the solving of block-Toeplitz linear systems by using Levinson--Durbin recursion~\cite{golub2013matrix,arushanian1983toeplitz} which has a complexity of $\mathcal{O}(N_t^2)$ in the time dimension.
Please note, that also so-called superfast algorithms exist~\cite{ammar1988superfast}, which further reduce the complexity to $\mathcal{O}(N_t \log N_t)$ and are faster in practice for $N_t > 256$.

A general spatiotemporal covariance matrix $\mat{\Sigma}$ for the background activity in a visual ERP task is shown in channel-prime form in~\Cref{fig:cov_free_params}.
To construct its stationary block-Toeplitz form, blocks sharing the same color are averaged, resulting in more robust block estimates per block-diagonal.
Depending on the block-diagonal only $N_t - d$ block matrices are available for averaging.
Note that if the number of epochs $N_e$ is small, this transformation can lead to matrices that are not positive definite any more---an issue which is resolved by taking the second assumption~\textbf{A2} into account.

To incorporate \textbf{A2}, we introduce a monotonically decreasing \emph{tapering} function~\cite{xiao2012covariance} $\tau(d)$. Using $\tau(d) \coloneqq 1-{|d|}/{N_t}$, a tapered channel-prime covariance matrix is obtained by
\begin{align*}
	\mat{C}_{i, j}^\text{tap} \stackrel{\text{\textbf{A2}}}{\coloneqq} \tau(i-j) \cdot \mat{C}_{i,j}.
\end{align*}
Combining both assumptions, we finally obtain
\begin{align*}
	\mat{\dot{\Sigma}} \stackrel{\text{\textbf{A1, A2}}}{\coloneqq} [\tau (i-j) \mat{C}_{\Delta t = i - j}]_{1 \leq i \leq N_t, 1 \leq j \leq N_t},
\end{align*}
which is a channel-prime block-Toeplitz covariance matrix with both assumptions incorporated.
Note that by averaging the block-diagonals and then applying the linear tapering function, the resulting block matrices are the sum along the block-diagonal with an additional scaling by $1/N_t$.

To facilitate this process in software, we published an open source package that builds on NumPy~\cite{harris2020array} and exposes all methods needed for working with block matrices with two domains to implement our approach.\footnote{\url{https://github.com/jsosulski/blockmatrix}}
As shrinkage regularization improves matrix conditioning and helps with solving the linear system, we apply our assumptions on top of a shrinkage regularized covariance matrix and term the resulting LDA `ToeplitzLDA'.

\subsection{Benchmark: Binary Classification of ERP Epochs}
\label{sec:benchmark_bin}

Across the datasets in~\Cref{sec:datasets} we apply ToeplitzLDA and compare it with an LDA that uses only shrinkage regularization (sLDA).
Additionally, we use the publicly available implementation of time-decoupled LDA~\cite{sosulski2021improving} (TDLDA)---a different approach to improve the covariance estimate---and a logistic regression classifier making use of Riemannian geometry~\cite{barachant2014plugplay} for comparison.
Contrary to an LDA classifier, the latter augments each ERP epoch with class templates and calculates a feature covariance matrix for every epoch, which is classified using logistic regression in the tangent space~\cite{kolkhorst2018guess}.
Note that the covariance matrices we talk about in this paper are very different as they represent EEG background activity, artifacts and other noise instead of features.

In most datasets six stimuli form a group, e.g., visual spellers with six rows/columns or auditory oddball paradigms with 1:5 target/non-target ratios.
Therefore we use $[6, 12, 24, 48, 96, 192, 384]$ epochs as sizes for the training subsets (if supported by the size of the dataset) and draw them randomly seven times.
An additional setting uses all epochs provided in the training set, the number of which varies between 300 and 7782, depending on the dataset.
On each subset, all classifiers are trained and then evaluated on $\datVal$.
We report average AUC performance for each subset size and each subject in every dataset, i.e., classification results are averaged across the randomly drawn subset and (if applicable) across multiple sessions of a subject.

Many BCIs use an aggregate of binary classification outputs obtained for several stimuli to solve a multi-class problem, e.g., in a visual ERP speller application the subject can choose between many different letters/symbols.
Due to this aggregation, even modest improvements in binary classification performance can lead to strong improvements of the application-level performance.

\subsection{Improving a Visual ERP Speller Application}
\label{sec:method_usup}

Most BCI setups consist of two phases: In the calibration phase labeled data is recorded, e.g., by instructing the user which symbol to attend to.
In the following productive/online phase, the user can actually operate the application according to his intention with the support of the trained classifier.
The usability of a BCI improves with shorter calibration and with quicker and more reliable detection of the user's intention, e.g., the desired symbol.
Unsupervised classifiers have been proposed that build upon either expectation maximization~\cite{kindermans2012bayesian}, learning from label proportions~\cite{quadrianto2009estimating,hubner2017learning} which makes use of known label proportions in the stimuli to estimate the class means, or a combination thereof~\cite{hubner2018unsupervised}. In visual speller BCIs their use allowed to eliminate the calibration phase entirely.

To evaluate the benefit of ToeplitzLDA in an actual BCI application, we used both the V\_LLP\_A and the V\_LLP\_B datasets, which had been recorded under a modified visual ERP paradigm that made learning from label proportions possible.

In these datasets, the subjects were instructed what to spell, however, the underlying LDA classifier did not receive this information but was trained unsupervisedly.
One trial corresponds to the spelling of one letter and consisted of 68 stimuli. The interval between the onset of two stimuli was 250\,ms such that one trial took approximately 20 seconds.

In our offline replay, we re-train the LDA classifier after each trial on the data of all past trials and---as we use an unsupervised classifier---also the current trial.
Only then the updated classifier is employed to predict the letter of the current trial by summing the classifier outputs separately for all possible letters and selecting the one with the maximal value.

In each of three blocks, a subject repeated the spelling of a sentence while the classifier was trained from scratch each block.
Note that in V\_LLP\_A a 63 letter sentence was spelled per block, whereas in V\_LLP\_B, only a 35 letter sentence was spelled per block.

While the class means can be estimated using learning from label proportions, we cannot calculate the within-class covariance matrix as label information would be necessary for this.
However, as shown by Hübner~\citeyear{hubner2020from} for two class problems, the global covariance can be used instead which does not require label information. Hübner showed, that the resulting weight vector defining LDA's projection has a different scaling but points into the same direction.

As the authors of the original publication used an sLDA classifier, we will compare the spelling performance of the sLDA with our ToeplitzLDA classifier.
To reproduce the results as closely as possible, we use the same bandpass filter ($[0.5, 8]$\,Hz) as in the original publication.
Regarding feature selection, the authors hand-picked $N_t=6$ differently sized time intervals and averaged the measurements inside the intervals of each epoch before passing it to the classifier.
Additionally, the authors corrected for baseline drifts---a frequently used technique for ERP analysis---by subtracting the average value of the baseline interval $[\shortminus0.2, 0]$\,s from each epoch, and omitted two artifact-prone frontal EEG channels.
Even though baseline correction and using differently sized time interval averages violate the assumptions we make in~\Cref{sec:assumptions} we still evaluate the performance of ToeplitzLDA in this setting as a robustness check.

To evaluate performance when our assumptions are not violated, we also evaluate a very basic feature selection settings, i.e., keeping all channels, applying no baseline correction and using all time samples in $[0.05, 0.70)$\,s relative to the stimulus onset.
In this setting, the number of time features $N_t$ is determined by the sampling rate. While a sampling rate of $20$\,Hz should reasonably contain all information for $8$\,Hz low-pass filtered data, we additionally evaluate the sampling rates $40, 100$ and $200$\,Hz. While these variants should not add new information, they shall reveal the impact of increasing time dimensionality on both, sLDA and ToeplitzLDA.
An implementation of ToeplitzLDA and the code to reproduce our results on the unsupervised speller paradigm can be found in our code repository\footnote{\url{https://github.com/jsosulski/toeplitzlda}}.

\section{Results}
\label{sec:results}

\subsection{Block-Toeplitz Structure in Visual ERPs}

\begin{figure}[t]
	\centering
	\includegraphics[width=\columnwidth]{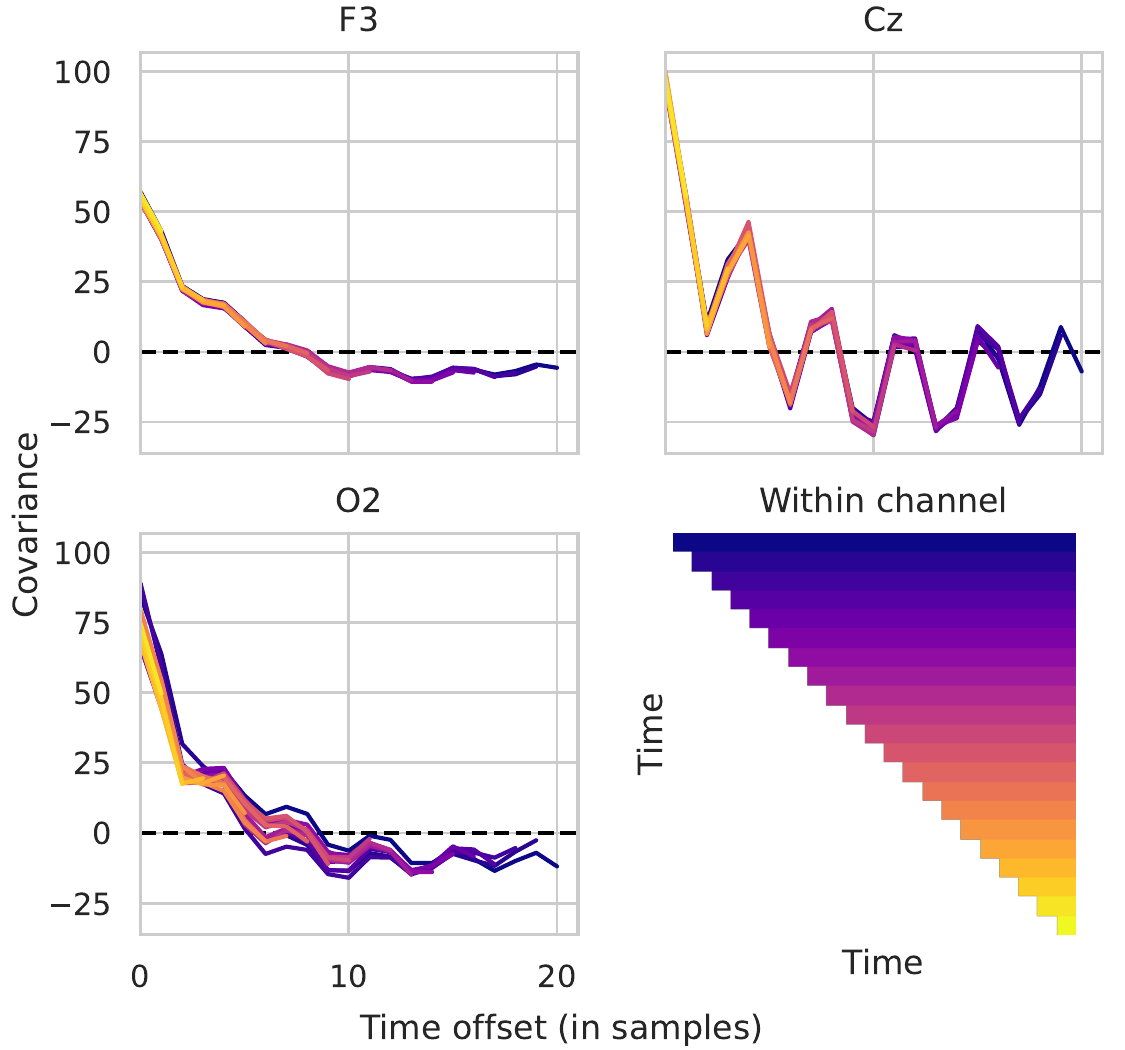}
	\caption{Temporal covariance within different channels of the EEG of subject 6 in the visual V\_LLP\_A dataset estimated from 3896 epochs. Data was bandpass filtered to $[0.5, 16]$\,Hz, transformed into epochs using the  $[0.1, 0.6)$\,s interval relative to each stimulus, and resampled to $40$\,Hz sampling rate. The legend in the lower right corner helps to interpret the shown temporal covariances: Dark blue entries relate time samples early in the interval with all later samples, while covariances between late time samples and all even later samples are indicated by bright yellow entries.}
	\label{fig:assumption}
\end{figure}

The temporal covariance within three different channels, estimated during a visual paradigm is depicted in \Cref{fig:assumption}.
If the data was perfectly stationary, the temporal covariance of each row would be overlapping.
This is approximately the case for channels F3 and Cz, but less so for channel O2.
Our method imposes the stationarity assumption on the covariance matrix, which corresponds to averaging the different temporal covariance curves per channel.

\subsection{Improved Binary Classification Performance on the ERP Benchmark}
\label{sec:results_bench}

For all LDA variants, using an ERP interval of $[0.1, 0.6)$\,s yielded the best performance on average across all datasets when using the complete training data.
The hand-picked intervals degrade in performance when they have to be generalized over various different datasets.
The Riemannian classification method performed best when applied upon six xDAWN components and when both target and non-target ERP templates were used.

\begin{figure}[t]
	\centering
	\includegraphics[width=\columnwidth]{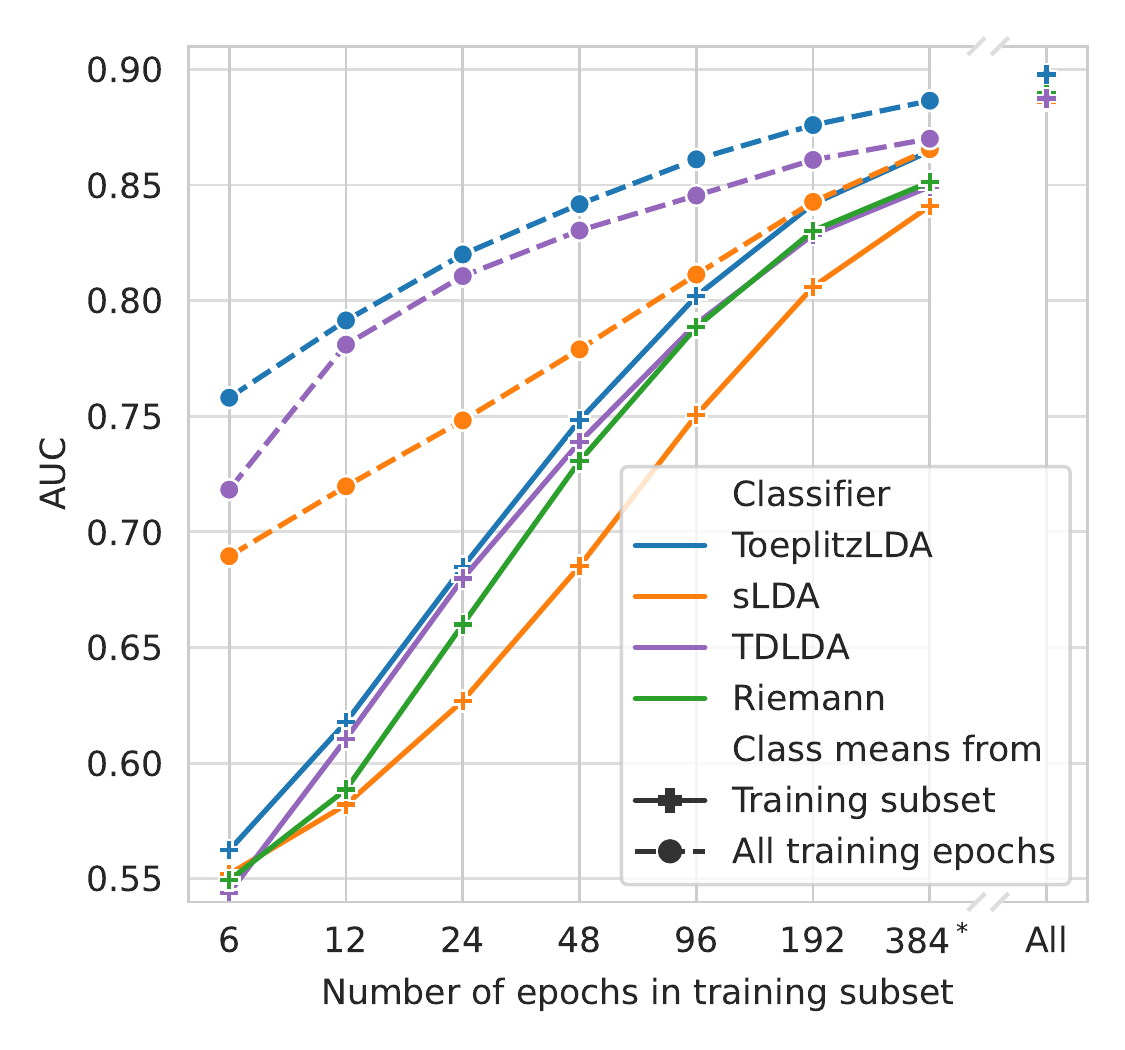}
	\caption{Classifier performance in AUC dependent on training subset size and classification approach. Solid lines with plus markers show the real-world performance, where solely the training subset has been used. For comparison, the dashed lines with circle markers indicate LDA variants, for which the best possible class mean information has been calculated from the whole training set and only the covariance matrix was estimated from smaller subsets of varying sizes. Generally, reported averages were obtained by first averaging all sessions of a subject, and then averaging across all subjects. Note that a subset size of 384 was supported by only 10 out of 13 datasets.}
	\label{fig:benchmark_global_average}
\end{figure}

In~\Cref{fig:benchmark_global_average}, the average AUC across all subjects and all datasets is shown.
For moderately sized training data subsets, TDLDA is close to ToeplitzLDA, however, when training data grows, ToeplitzLDA has an advantage.
Note that the Riemannian classifier performs markedly better than the state of the art sLDA, but is worse than the further optimized LDA variants TDLDA and ToeplitzLDA.
For sLDA to achieve the same AUC as ToeplitzLDA, it requires a nearly twice as much training data.
Using a two-sided paired Wilcoxon rank sum test with Holm-Bonferroni correction for 24 tests, ToeplitzLDA is significantly better than all the other methods, with the sole exception of TDLDA for a subset size of 24 epochs.
For a per dataset analysis see~\Cref{sec:supp:benchmark_results}.

The difference between ToeplitzLDA and sLDA is even more pronounced when considering the hypothetical case where the class mean estimates are known a priori, but the covariance is not.
While this setting is mostly of theoretic interest, it emphasizes that even with a very accurate mean estimation, performance of LDA classifiers depend strongly on the estimation of the covariance matrix.
In this setting, sLDA needs somewhere between four to eight times as much data to reach the AUC of the ToeplitzLDA.

\begin{figure}[t]
	\centering
	\includegraphics[width=\columnwidth]{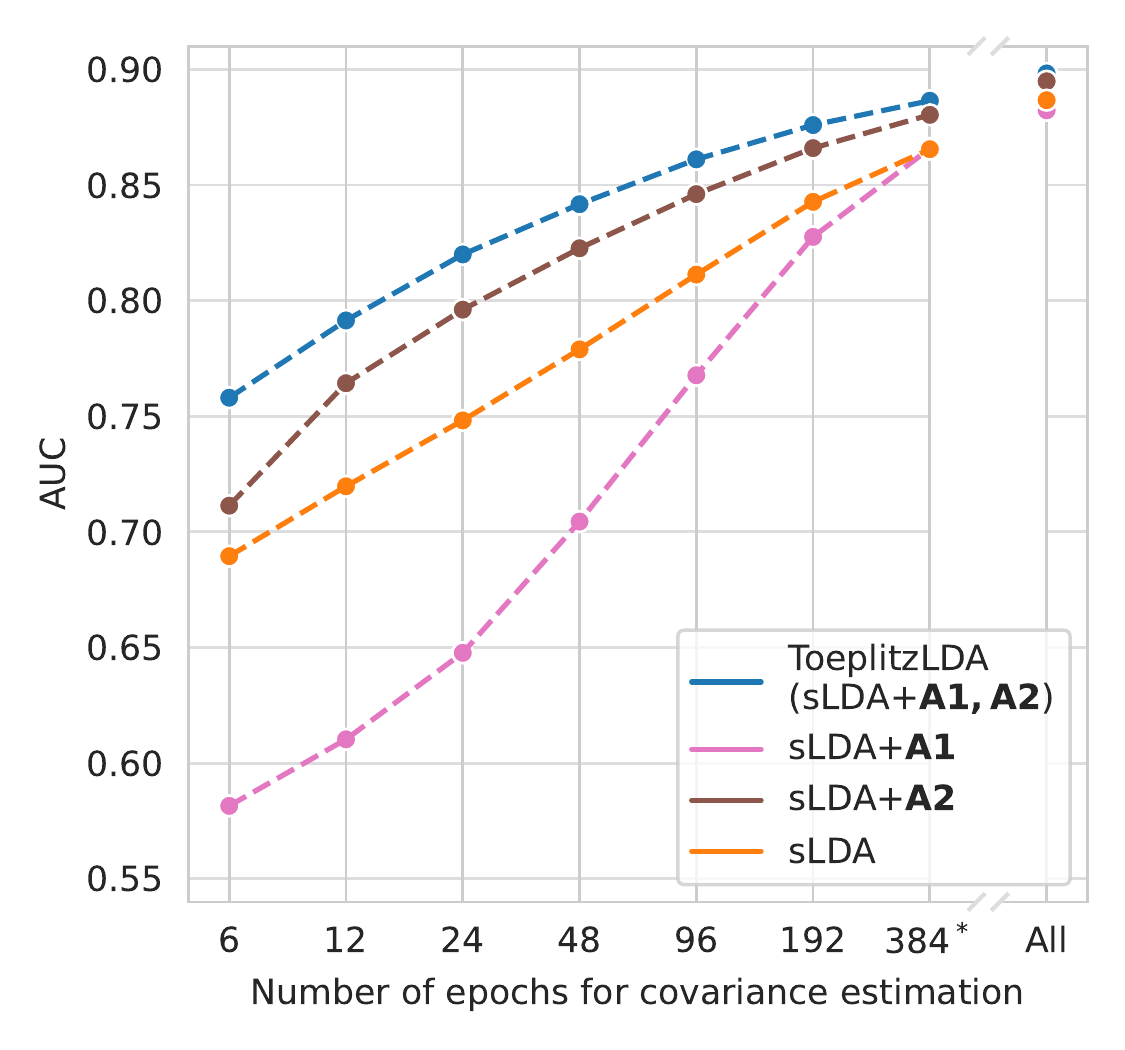}
	\caption{Average classifier performance in AUC in the setting where only the covariance is estimated from the subset and the mean estimation uses all epochs from the whole training set. Subsets of 384 epochs could be reached only for 10 out of 13 datasets.}
	\label{fig:benchmark_influence_of_assumptions}
\end{figure}

Finally, in~\Cref{fig:benchmark_influence_of_assumptions} the influence of each assumption upon the classification performance is shown for a hypothetical setting where the mean information is known.
Applying \textbf{A2} only, i.e., a linear tapering on the covariance matrix, already leads to better performances compared to sLDA.
Contrary, enforcing time stationarity only (\textbf{A1}) by using the averaging method decreases performance markedly.
This can be explained by the different number of epochs available for very small time distances compared to very large time distances (see~\Cref{sec:assumptions}).
Combining both assumptions eliminates the issue with non-positive definiteness of the transformed matrices and achieves the best classification performances.

\subsection{Effect on an Unsupervised Visual ERP Speller Application}
\label{sec:results_appl}

In contrast to the binary classification benchmark reported so far, classifiers in the unsupervised ERP speller setting had to cope with approximate class means (provided by learning from label proportions) and the global covariance matrix, as labeled data was not made available.

\begin{figure}
	\centering
	\includegraphics[width=\columnwidth]{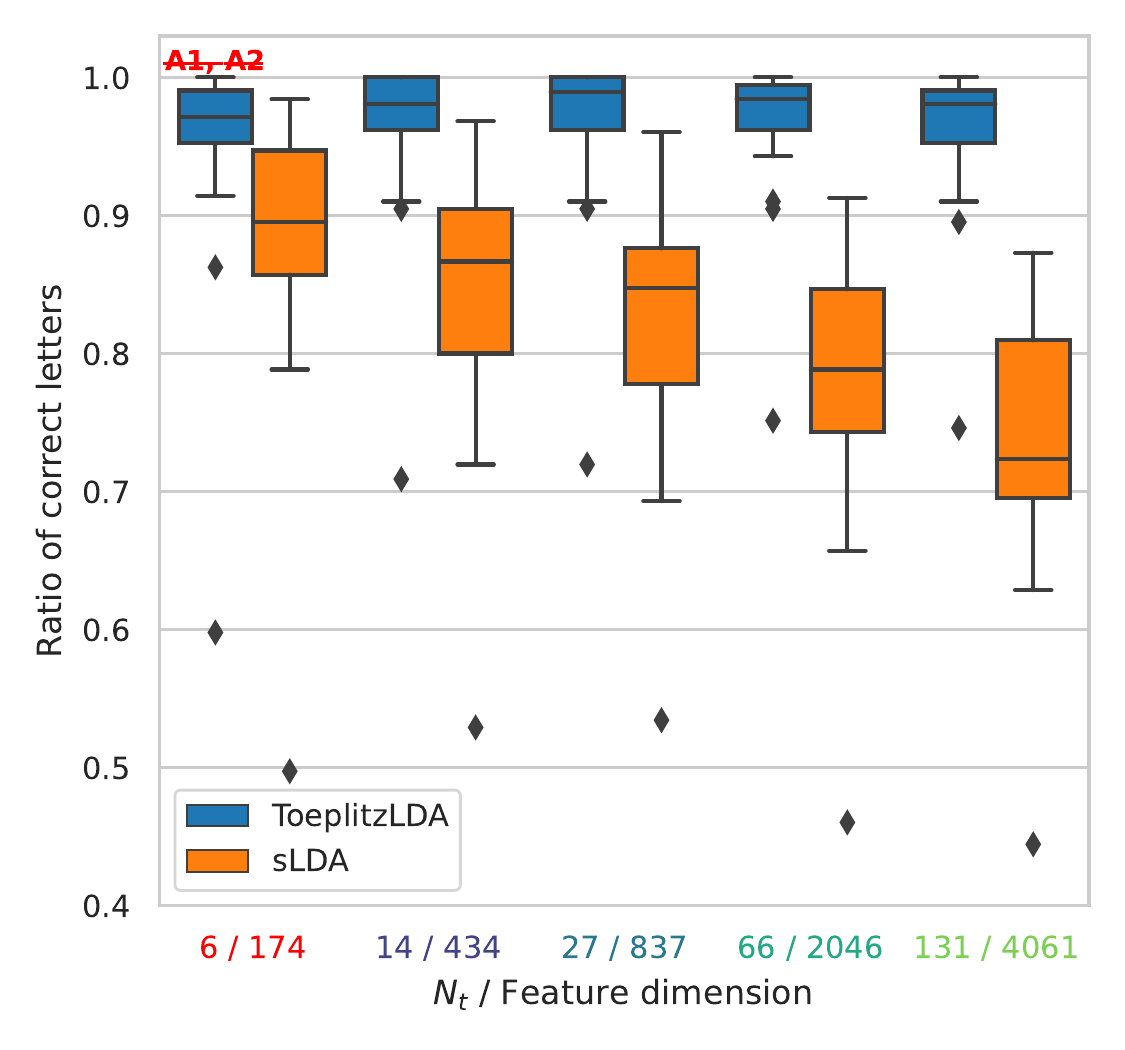}
	\caption{Ratio of correctly spelled letters (all subjects, LLP datasets). For $N_t=6$ the preprocessing violated two assumptions of ToeplitzLDA. Feature dimensionality is the number of time samples multiplied by the number of channels. The best median performance for sLDA is $0.895$, obtained using $N_t=6$. ToeplitzLDA reaches the median AUC $0.989$ for $N_t=27$.}
	\label{fig:speller_correct_rate}
\end{figure}

The results of applying both LDA classifiers to the data of 25 subjects are shown in~\Cref{fig:speller_correct_rate}.
Spelling errors were reduced on average by 81\,\% across all 25 subjects (see~\Cref{sec:supp:appl_results}).
Interestingly, even when assumptions are explicitly violated in the setting $N_t=6$, the ToeplitzLDA classifier performs better than sLDA.
For larger feature dimensions performance of sLDA degrades quickly, whereas this effect is minimal for ToeplitzLDA.
Please note the almost stable performance of ToeplitzLDA for thousands of features.

\begin{figure}
	\centering
	\includegraphics[width=\columnwidth]{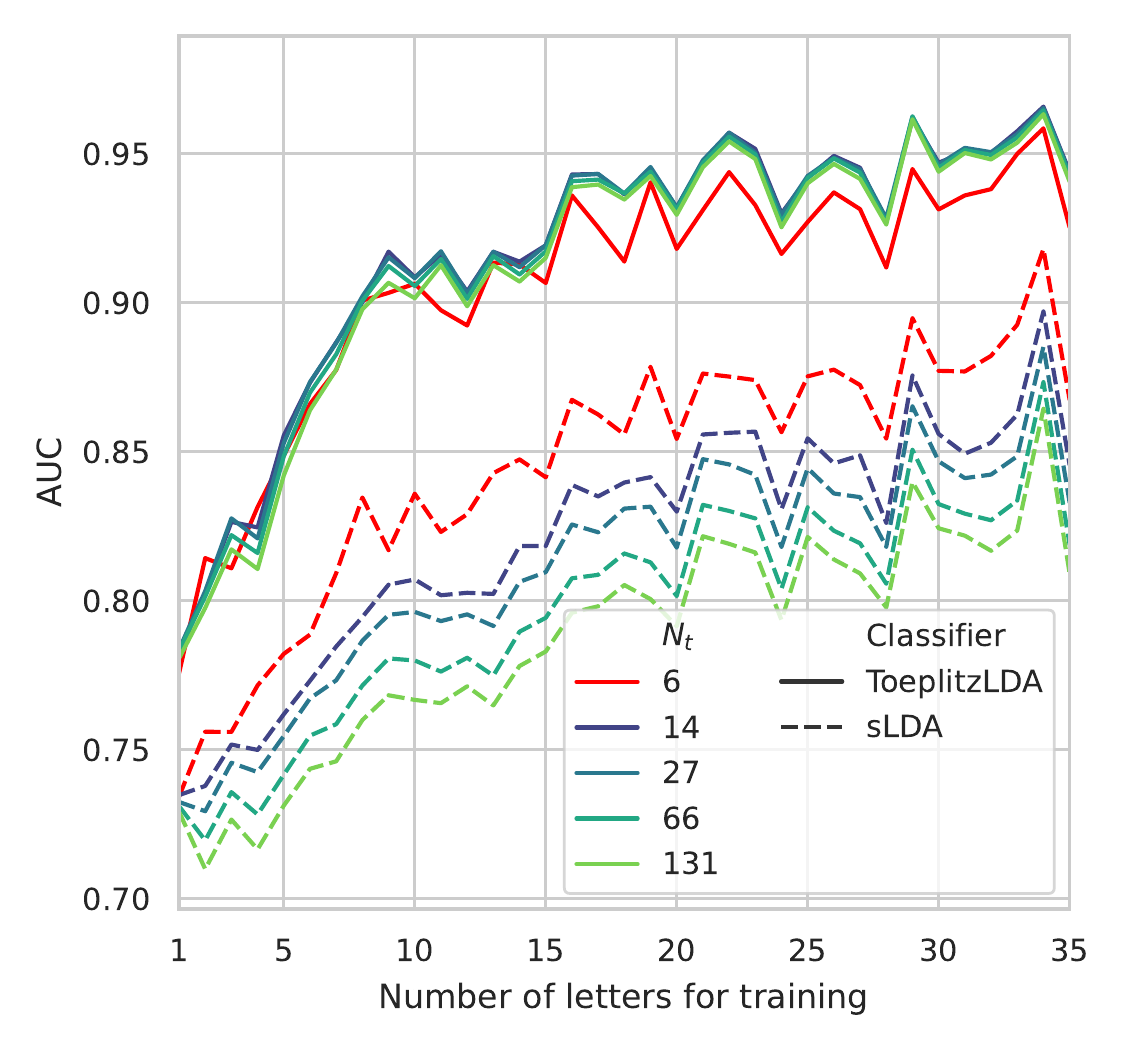}
	\caption{Averaged learning curves of sLDA and ToeplitzLDA for the first 35 letters (=2380 epochs), averaged over three blocks and 25 subjects. Classifiers were trained on all letters up to and including the one that was evaluated on (see~\Cref{sec:method_usup}) and updated after each letter. Red lines indicate scenarios violating assumptions \textbf{A1} and \textbf{A2}.}
	\label{fig:speller_auc_learning_curve}
\end{figure}

As shown in~\Cref{fig:speller_auc_learning_curve}, the violation of the assumptions of ToeplitzLDA under a typical preprocessing for $N_t=6$ impacts performance only when more data is available.
These results also emphasize that sLDA requires careful consideration which time interval to use.
Interestingly, the performance advantage of ToeplitzLDA is more pronounced on this LLP data compared to the benchmark, were the true class labels were used.
To reach 0.80, 0.85 and 0.90 AUC, ToeplitzLDA needs approximately 2, 5, 8 letters respectively, whereas sLDA requires 7, 16, 33 letters.

\section{Discussion}
\label{sec:discussion}

Making use of domain specific assumptions for ERP-based BCI improves the performance of an sLDA classifier by up to 6 AUC points across 213 subjects from 13 different ERP datasets.
The proposed ToeplitzLDA additionally obtains better AUC values compared to other state of the art classification methods in BCI.
This shows that the relatively simple but still popular~\cite{lotte2018review} LDA classifier can be brought up to speed  by using this novel covariance estimation.
Reducing the free parameter count makes the covariance estimation more efficient, which may explain ToeplitzLDA's performance benefits especially on small data sets.
As improvements were observed even for large training data, the enforced assumptions may also prevent overfitting to artifacts, which otherwise cause seemingly non-stationary covariance matrix estimates.
While the linear tapering is already successful, other tapering methods should be evaluated in future work.

In an unsupervised visual speller BCI application, the ToeplitzLDA reduces the median spelling error from $10.5$\,\% to $1.1$\,\% compared to the originally used sLDA.
Aside from the obvious benefit of making fewer errors, BCI users perceive less frustration~\cite{hougaard2021who} when the error rate is small, and motivated engagement with the BCI may improve signal quality~\cite{kleih2010motivation}.
As our method reduces errors especially in the first few minutes of usage, the BCI user faces a productive system from the start on.

By enforcing the block-Toeplitz structure, we also reduce time and memory complexity of LDA training.
This is especially useful in applications requiring frequent re-trainings like the unsupervised visual speller, which estimates a new weight vector after every spelled letter.
Note that in our implementation we still store the full covariance matrix to facilitate the averaging process of the block-diagonals, however, this could be avoided by calculating incremental updates of the block-matrices.

Using a structured covariance matrix has been explored earlier in the domain of EEG signal modeling and classification.
This comprises modeling the EEG's spatiotemporal covariance as a Kronecker product of parameterized temporal and spatial covariance matrices~\cite{huizenga2002spatiotemporal} or applying the Kronecker assumption on the generating sources~\cite{hashemi2021efficient}, which improves source estimation compared to using an ordinarily estimated covariance.
Estimating respective temporal and spatial covariance matrices from the data, but shrinking them towards different targets~\cite{beltrachini2013shrinkage} led to a reduced distance between estimated and true covariance matrix, if the power values of EEG channels are similar.
Also using a Kronecker product structure for the covariance matrix of a linear discriminant analysis, Gonzalez-Navarro et al.~(\citeyear{gonzalez-navarro2017spatio-temporal}) report an increased performance for 12 subjects and small training data sets.
However, the authors used a large number of time samples ($N_t = 64$), which is a difficult and unusual feature setting for an sLDA (see~\Cref{sec:results_appl}).
A Kronecker product structured covariance in the sensor space is most suitable, if (up to scaling) a single temporal covariance curve matches the covariance observed within each channel.
Given the observations in~\Cref{fig:assumption} this seems too restrictive for ERP classification with LDA, whereas our proposed block-Toeplitz structure is very close to the data.
We also found that slight violations of the stationarity assumptions, i.e., by averaging non-regular ERP intervals or baseline correction, seems to be detrimental only, when at least data of 10 letters (680 epochs) has been recorded.
This may indicate why a Kronecker product structured covariance can still be successful with fewer training epochs.
Currently, many ERP-based BCIs still average ERP signals in specific time intervals to reduce time dimensionality and thus to make sLDA feasible.
If the intervals are chosen well, this approach works.
Knowing which intervals to use, however, requires expert knowledge and optimal intervals may vary between sessions, subjects and paradigms.
Using ToeplitzLDA makes this step obsolete, as every time sample in a typical ERP interval can and should be used.
Practically, we found our method performs well even when artifacts or time-locked noise violate the stationarity assumption.
However, employing automatic artifact removal techniques \cite{islam2016methods} may align the signal even better with our assumptions and should be evaluated in further work.

\section{Conclusion}
\label{sec:conclusion}

We proposed to exploit inherent structure in ERP electroencephalogram data to greatly improve the classification performance of LDA across many different ERP datasets.
In an unsupervised visual speller BCI, our method reduces the error rate by 81\,\% and could even allow seven subjects to spell perfectly starting from the very first letter.
Additionally, using the proposed ToeplitzLDA lowers the complexity of LDA training considerably, reducing the needed computational resources.

\section*{Acknowledgements}

Our work was supported by the German Research Foundation project SuitAble (DFG, grant number 387670982) and by the Federal Ministry of Education and Research (BMBF, grant number 16SV8012). The authors would also like to acknowledge support by the state of Baden-W\"{u}rttemberg, Germany, through bwHPC and the German Research Foundation (DFG, INST 39/963-1 FUGG).

\bibliography{bib/master}

\begin{thebibliography}{50}
\providecommand{\natexlab}[1]{#1}
\providecommand{\url}[1]{\texttt{#1}}
\expandafter\ifx\csname urlstyle\endcsname\relax
  \providecommand{\doi}[1]{doi: #1}\else
  \providecommand{\doi}{doi: \begingroup \urlstyle{rm}\Url}\fi

\bibitem[Ammar \& Gragg(1988)Ammar and Gragg]{ammar1988superfast}
Ammar, G.~S. and Gragg, W.~B.
\newblock Superfast solution of real positive definite {Toeplitz} systems.
\newblock \emph{SIAM Journal on Matrix Analysis and Applications}, 9\penalty0
  (1):\penalty0 61--76, 1988.

\bibitem[Ang \& Guan(2013)Ang and Guan]{ang2013brain-computer}
Ang, K.~K. and Guan, C.
\newblock Brain-computer interface in stroke rehabilitation.
\newblock 2013.

\bibitem[Aric{\`o} et~al.(2014)Aric{\`o}, Aloise, Schettini, Salinari, Mattia,
  and Cincotti]{arico2014influence}
Aric{\`o}, P., Aloise, F., Schettini, F., Salinari, S., Mattia, D., and
  Cincotti, F.
\newblock Influence of {P300} latency jitter on event related potential-based
  brain--computer interface performance.
\newblock \emph{Journal of neural engineering}, 11\penalty0 (3):\penalty0
  035008, 2014.

\bibitem[Arushanian et~al.(1983)Arushanian, Boyle, Dritz, Voevodin, Garbow,
  Tyrtyshnikov, Cowell, and Samarin]{arushanian1983toeplitz}
Arushanian, O., Boyle, J., Dritz, K., Voevodin, V., Garbow, B., Tyrtyshnikov,
  E., Cowell, W., and Samarin, M.
\newblock The {Toeplitz} package users' guide.
\newblock Technical report, CM-P00069203, 1983.

\bibitem[Barachant \& Congedo(2014)Barachant and
  Congedo]{barachant2014plugplay}
Barachant, A. and Congedo, M.
\newblock A plug\&play {P300} {BCI} using information geometry.
\newblock \emph{arXiv preprint arXiv:1409.0107}, 2014.

\bibitem[Beltrachini et~al.(2013)Beltrachini, von Ellenrieder, and
  Muravchik]{beltrachini2013shrinkage}
Beltrachini, L., von Ellenrieder, N., and Muravchik, C.~H.
\newblock Shrinkage approach for spatiotemporal {EEG} covariance matrix
  estimation.
\newblock \emph{IEEE Transactions on Signal Processing}, 61\penalty0
  (7):\penalty0 1797--1808, 2013.

\bibitem[Bishop(2006)]{bishop2006linear}
Bishop, C.~M.
\newblock Linear models for classification.
\newblock In \emph{Pattern recognition and machine learning}, chapter~4, pp.\
  179--220. Springer, 2006.

\bibitem[Blankertz et~al.(2011)Blankertz, Lemm, Treder, Haufe, and
  M{\"u}ller]{blankertz2011single-trial}
Blankertz, B., Lemm, S., Treder, M., Haufe, S., and M{\"u}ller, K.-R.
\newblock Single-trial analysis and classification of {ERP} components---a
  tutorial.
\newblock \emph{NeuroImage}, 56\penalty0 (2):\penalty0 814--825, 2011.

\bibitem[Chen et~al.(2013)Chen, Xu, Wu, et~al.]{chen2013covariance}
Chen, X., Xu, M., Wu, W.~B., et~al.
\newblock Covariance and precision matrix estimation for high-dimensional time
  series.
\newblock \emph{Annals of Statistics}, 41\penalty0 (6):\penalty0 2994--3021,
  2013.

\bibitem[Cohen \& Sances(1977)Cohen and Sances]{cohen1977stationarity}
Cohen, B.~A. and Sances, A.
\newblock Stationarity of the human electroencephalogram.
\newblock \emph{Medical and Biological Engineering and Computing}, 15\penalty0
  (5):\penalty0 513--518, 1977.

\bibitem[Denzer(2016)]{denzer2016influence}
Denzer, S.
\newblock Influence of spatial word presentation in an auditory {ERP} paradigm
  for brain-computer interfaces.
\newblock Master's thesis, 2016.

\bibitem[Farwell \& Donchin(1988)Farwell and Donchin]{farwell1988talking}
Farwell, L.~A. and Donchin, E.
\newblock Talking off the top of your head: toward a mental prosthesis
  utilizing event-related brain potentials.
\newblock \emph{Electroencephalography and Clinical Neurophysiology},
  70\penalty0 (6):\penalty0 510--523, 1988.

\bibitem[Fern{\'a}ndez-Rodr{\'\i}guez et~al.(2016)Fern{\'a}ndez-Rodr{\'\i}guez,
  Velasco-{\'A}lvarez, and Ron-Angevin]{fernandez-rodrguez2016review}
Fern{\'a}ndez-Rodr{\'\i}guez, {\'A}., Velasco-{\'A}lvarez, F., and Ron-Angevin,
  R.
\newblock Review of real brain-controlled wheelchairs.
\newblock \emph{Journal of neural engineering}, 13\penalty0 (6):\penalty0
  061001, 2016.

\bibitem[Furrer \& Bengtsson(2007)Furrer and Bengtsson]{furrer2007estimation}
Furrer, R. and Bengtsson, T.
\newblock Estimation of high-dimensional prior and posterior covariance
  matrices in {Kalman} filter variants.
\newblock \emph{Journal of Multivariate Analysis}, 98\penalty0 (2):\penalty0
  227--255, 2007.

\bibitem[Golub \& Van~Loan(2013)Golub and Van~Loan]{golub2013matrix}
Golub, G. and Van~Loan, C.
\newblock \emph{Matrix Computations}.
\newblock Johns Hopkins Studies in the Mathematical Sciences. Johns Hopkins
  University Press, 2013.
\newblock ISBN 9781421407944.

\bibitem[Gonzalez-Navarro et~al.(2017)Gonzalez-Navarro, Moghadamfalahi,
  Ak{\c{c}}akaya, and Erdogmus]{gonzalez-navarro2017spatio-temporal}
Gonzalez-Navarro, P., Moghadamfalahi, M., Ak{\c{c}}akaya, M., and Erdogmus, D.
\newblock Spatio-temporal {EEG} models for brain interfaces.
\newblock \emph{Signal processing}, 131:\penalty0 333--343, 2017.

\bibitem[Gruenwald et~al.(2019)Gruenwald, Znobishchev, Kapeller, Kamada,
  Scharinger, and Guger]{gruenwald2019time-variant}
Gruenwald, J., Znobishchev, A., Kapeller, C., Kamada, K., Scharinger, J., and
  Guger, C.
\newblock Time-variant linear discriminant analysis improves hand gesture and
  finger movement decoding for invasive brain-computer interfaces.
\newblock \emph{Frontiers in neuroscience}, 13:\penalty0 901, 2019.

\bibitem[Guger et~al.(2009)Guger, Daban, Sellers, Holzner, Krausz, Carabalona,
  Gramatica, and Edlinger]{guger2009how}
Guger, C., Daban, S., Sellers, E., Holzner, C., Krausz, G., Carabalona, R.,
  Gramatica, F., and Edlinger, G.
\newblock How many people are able to control a {P300}-based brain--computer
  interface ({BCI})?
\newblock \emph{Neuroscience letters}, 462\penalty0 (1):\penalty0 94--98, 2009.

\bibitem[Harris et~al.(2020)Harris, Millman, van~der Walt, Gommers, Virtanen,
  Cournapeau, Wieser, Taylor, Berg, Smith, Kern, Picus, Hoyer, van Kerkwijk,
  Brett, Haldane, del R{\'{i}}o, Wiebe, Peterson, G{\'{e}}rard-Marchant,
  Sheppard, Reddy, Weckesser, Abbasi, Gohlke, and Oliphant]{harris2020array}
Harris, C.~R., Millman, K.~J., van~der Walt, S.~J., Gommers, R., Virtanen, P.,
  Cournapeau, D., Wieser, E., Taylor, J., Berg, S., Smith, N.~J., Kern, R.,
  Picus, M., Hoyer, S., van Kerkwijk, M.~H., Brett, M., Haldane, A., del
  R{\'{i}}o, J.~F., Wiebe, M., Peterson, P., G{\'{e}}rard-Marchant, P.,
  Sheppard, K., Reddy, T., Weckesser, W., Abbasi, H., Gohlke, C., and Oliphant,
  T.~E.
\newblock Array programming with {NumPy}.
\newblock \emph{Nature}, 585\penalty0 (7825):\penalty0 357--362, September
  2020.
\newblock \doi{10.1038/s41586-020-2649-2}.
\newblock URL \url{https://doi.org/10.1038/s41586-020-2649-2}.

\bibitem[Hashemi et~al.(2021)Hashemi, Gao, Cai, Ghosh, M{\"u}ller, Nagarajan,
  and Haufe]{hashemi2021efficient}
Hashemi, A., Gao, Y., Cai, C., Ghosh, S., M{\"u}ller, K.-R., Nagarajan, S., and
  Haufe, S.
\newblock Efficient hierarchical {Bayesian} inference for spatio-temporal
  regression models in neuroimaging.
\newblock \emph{Advances in Neural Information Processing Systems}, 34, 2021.

\bibitem[Hoffmann et~al.(2008)Hoffmann, Vesin, Ebrahimi, and
  Diserens]{hoffmann2008efficient}
Hoffmann, U., Vesin, J.-M., Ebrahimi, T., and Diserens, K.
\newblock An efficient {P300}-based brain--computer interface for disabled
  subjects.
\newblock \emph{Journal of neuroscience methods}, 167\penalty0 (1):\penalty0
  115--125, 2008.

\bibitem[H{\"o}hne et~al.(2012)H{\"o}hne, Krenzlin, D{\"a}hne, and
  Tangermann]{hohne2012natural}
H{\"o}hne, J., Krenzlin, K., D{\"a}hne, S., and Tangermann, M.
\newblock Natural stimuli improve auditory {BCIs} with respect to ergonomics
  and performance.
\newblock \emph{Journal of Neural Engineering}, 9\penalty0 (4):\penalty0
  045003, 2012.

\bibitem[Hougaard et~al.(2021)Hougaard, Rossau, Czapla, Miko, Skammelsen,
  Knoche, and Jochumsen]{hougaard2021who}
Hougaard, B.~I., Rossau, I.~G., Czapla, J.~J., Miko, M.~A., Skammelsen, R.~B.,
  Knoche, H., and Jochumsen, M.
\newblock Who willed it? {Decreasing} frustration by manipulating perceived
  control through fabricated input for stroke rehabilitation {BCI} games.
\newblock In \emph{CHI Play'21}. 2021.

\bibitem[H{\"u}bner(2020)]{hubner2020from}
H{\"u}bner, D.
\newblock \emph{From Supervised to Unsupervised Machine Learning Methods for
  Brain-Computer Interfaces and Their Application in Language Rehabilitation}.
\newblock PhD thesis, University of Freiburg, 2020.

\bibitem[H{\"u}bner et~al.(2017)H{\"u}bner, Verhoeven, Schmid, M{\"u}ller,
  Tangermann, and Kindermans]{hubner2017learning}
H{\"u}bner, D., Verhoeven, T., Schmid, K., M{\"u}ller, K.-R., Tangermann, M.,
  and Kindermans, P.-J.
\newblock Learning from label proportions in brain-computer interfaces: online
  unsupervised learning with guarantees.
\newblock \emph{PloS one}, 12\penalty0 (4), 2017.

\bibitem[H{\"u}bner et~al.(2018)H{\"u}bner, Verhoeven, M{\"u}ller, Kindermans,
  and Tangermann]{hubner2018unsupervised}
H{\"u}bner, D., Verhoeven, T., M{\"u}ller, K.-R., Kindermans, P.-J., and
  Tangermann, M.
\newblock Unsupervised learning for brain-computer interfaces based on
  event-related potentials: Review and online comparison.
\newblock \emph{IEEE Computational Intelligence Magazine}, 13\penalty0
  (2):\penalty0 66--77, 2018.

\bibitem[Huizenga et~al.(2002)Huizenga, De~Munck, Waldorp, and
  Grasman]{huizenga2002spatiotemporal}
Huizenga, H.~M., De~Munck, J.~C., Waldorp, L.~J., and Grasman, R.~P.
\newblock Spatiotemporal {EEG}/{MEG} source analysis based on a parametric
  noise covariance model.
\newblock \emph{IEEE Transactions on Biomedical Engineering}, 49\penalty0
  (6):\penalty0 533--539, 2002.

\bibitem[Islam et~al.(2016)Islam, Rastegarnia, and Yang]{islam2016methods}
Islam, M.~K., Rastegarnia, A., and Yang, Z.
\newblock Methods for artifact detection and removal from scalp {EEG}: A
  review.
\newblock \emph{Neurophysiologie Clinique/Clinical Neurophysiology},
  46\penalty0 (4-5):\penalty0 287--305, 2016.

\bibitem[Jayaram \& Barachant(2018)Jayaram and Barachant]{jayaram2018moabb}
Jayaram, V. and Barachant, A.
\newblock {MOABB}: trustworthy algorithm benchmarking for {BCIs}.
\newblock \emph{Journal of neural engineering}, 15\penalty0 (6):\penalty0
  066011, 2018.

\bibitem[Jayaram et~al.(2016)Jayaram, Alamgir, Altun, Scholkopf, and
  Grosse-Wentrup]{jayaram2016transfer}
Jayaram, V., Alamgir, M., Altun, Y., Scholkopf, B., and Grosse-Wentrup, M.
\newblock Transfer learning in brain-computer interfaces.
\newblock \emph{IEEE Computational Intelligence Magazine}, 11\penalty0
  (1):\penalty0 20--31, 2016.

\bibitem[Kindermans et~al.(2012)Kindermans, Verstraeten, and
  Schrauwen]{kindermans2012bayesian}
Kindermans, P.-J., Verstraeten, D., and Schrauwen, B.
\newblock A {Bayesian} model for exploiting application constraints to enable
  unsupervised training of a {P300}-based {BCI}.
\newblock \emph{PloS one}, 7\penalty0 (4):\penalty0 e33758, 2012.

\bibitem[Kleih et~al.(2010)Kleih, Nijboer, Halder, and
  K{\"u}bler]{kleih2010motivation}
Kleih, S., Nijboer, F., Halder, S., and K{\"u}bler, A.
\newblock Motivation modulates the {P300} amplitude during brain--computer
  interface use.
\newblock \emph{Clinical Neurophysiology}, 121\penalty0 (7):\penalty0
  1023--1031, 2010.

\bibitem[Kolkhorst et~al.(2018)Kolkhorst, Tangermann, and
  Burgard]{kolkhorst2018guess}
Kolkhorst, H., Tangermann, M., and Burgard, W.
\newblock Guess what {I} attend: Interface-free object selection using brain
  signals.
\newblock In \emph{2018 IEEE/RSJ International Conference on Intelligent Robots
  and Systems (IROS)}, pp.\  7111--7116. IEEE, 2018.

\bibitem[Ledoit \& Wolf(2004)Ledoit and Wolf]{ledoit2004well-conditioned}
Ledoit, O. and Wolf, M.
\newblock A well-conditioned estimator for large-dimensional covariance
  matrices.
\newblock \emph{Journal of multivariate analysis}, 88\penalty0 (2):\penalty0
  365--411, 2004.

\bibitem[Lee et~al.(2019)Lee, Kwon, Kim, Kim, Lee, Williamson, Fazli, and
  Lee]{lee2019eeg}
Lee, M.-H., Kwon, O.-Y., Kim, Y.-J., Kim, H.-K., Lee, Y.-E., Williamson, J.,
  Fazli, S., and Lee, S.-W.
\newblock {EEG} dataset and {OpenBMI} toolbox for three {BCI} paradigms: an
  investigation into {BCI} illiteracy.
\newblock \emph{GigaScience}, 8\penalty0 (5):\penalty0 giz002, 2019.

\bibitem[Lim et~al.(2012)Lim, Lee, Guan, Fung, Zhao, Teng, Zhang, and
  Krishnan]{lim2012brain-computer}
Lim, C.~G., Lee, T.~S., Guan, C., Fung, D. S.~S., Zhao, Y., Teng, S. S.~W.,
  Zhang, H., and Krishnan, K. R.~R.
\newblock A brain-computer interface based attention training program for
  treating attention deficit hyperactivity disorder.
\newblock \emph{PloS one}, 7\penalty0 (10):\penalty0 e46692, 2012.

\bibitem[Lin et~al.(2018)Lin, Zhang, Zeng, Tong, and Yan]{lin2018novel}
Lin, Z., Zhang, C., Zeng, Y., Tong, L., and Yan, B.
\newblock A novel {P300} {BCI} speller based on the triple {RSVP} paradigm.
\newblock \emph{Scientific reports}, 8\penalty0 (1):\penalty0 3350, 2018.

\bibitem[Lotte et~al.(2007)Lotte, Congedo, L{\'e}cuyer, Lamarche, and
  Arnaldi]{lotte2007review}
Lotte, F., Congedo, M., L{\'e}cuyer, A., Lamarche, F., and Arnaldi, B.
\newblock A review of classification algorithms for {EEG}-based brain--computer
  interfaces.
\newblock \emph{Journal of Neural Engineering}, 4\penalty0 (2):\penalty0 R1,
  2007.

\bibitem[Lotte et~al.(2018)Lotte, Bougrain, Cichocki, Clerc, Congedo,
  Rakotomamonjy, and Yger]{lotte2018review}
Lotte, F., Bougrain, L., Cichocki, A., Clerc, M., Congedo, M., Rakotomamonjy,
  A., and Yger, F.
\newblock A review of classification algorithms for {EEG}-based brain--computer
  interfaces: a 10 year update.
\newblock \emph{Journal of neural engineering}, 15\penalty0 (3):\penalty0
  031005, 2018.

\bibitem[Musso et~al.(2022)Musso, H\"{u}bner, Schwarzkopf, Bernodusson, LeVan,
  Weiller, and Tangermann]{musso2022aphasia}
Musso, M., H\"{u}bner, D., Schwarzkopf, S., Bernodusson, M., LeVan, P.,
  Weiller, C., and Tangermann, M.
\newblock Aphasia recovery by language training using a brain-computer
  interface -- a proof-of-concept study.
\newblock \emph{Brain Communications}, 2022.
\newblock \doi{10.1093/braincomms/fcac008}.

\bibitem[Nunez(2012)]{nunez2012electric}
Nunez, P.~L.
\newblock Electric and magnetic fields produced by the brain.
\newblock \emph{Brain-Computer Interfaces: Principles and Practice}, pp.\
  171--212, 2012.

\bibitem[Pourahmadi(2013)]{pourahmadi2013banding}
Pourahmadi, M.
\newblock \emph{Banding, Tapering and Thresholding}, volume 882, chapter~6,
  pp.\  141--151.
\newblock John Wiley \& Sons, 2013.

\bibitem[Quadrianto et~al.(2009)Quadrianto, Smola, Caetano, and
  Le]{quadrianto2009estimating}
Quadrianto, N., Smola, A.~J., Caetano, T.~S., and Le, Q.~V.
\newblock Estimating labels from label proportions.
\newblock \emph{Journal of Machine Learning Research}, 10\penalty0 (10), 2009.

\bibitem[Rivet et~al.(2009)Rivet, Souloumiac, Attina, and
  Gibert]{rivet2009xdawn}
Rivet, B., Souloumiac, A., Attina, V., and Gibert, G.
\newblock {xDAWN} algorithm to enhance evoked potentials: application to
  brain--computer interface.
\newblock \emph{IEEE Transactions on Biomedical Engineering}, 56\penalty0
  (8):\penalty0 2035--2043, 2009.

\bibitem[Schreuder et~al.(2010)Schreuder, Blankertz, and
  Tangermann]{schreuder2010new}
Schreuder, M., Blankertz, B., and Tangermann, M.
\newblock A new auditory multi-class brain-computer interface paradigm: Spatial
  hearing as an informative cue.
\newblock \emph{PLOS ONE}, 5\penalty0 (4):\penalty0 1--14, 04 2010.
\newblock \doi{10.1371/journal.pone.0009813}.
\newblock URL \url{https://doi.org/10.1371/journal.pone.0009813}.

\bibitem[Sellers et~al.(2012)Sellers, Arbel, and Donchin]{sellers2012bcis}
Sellers, E.~W., Arbel, Y., and Donchin, E.
\newblock {BCIs} that use {P300} event-related potentials.
\newblock In \emph{Brain-Computer Interfaces: Principles and Practice}, pp.\
  215. Oxford University Press, 2012.

\bibitem[Sosulski \& Tangermann(2019)Sosulski and
  Tangermann]{sosulski2019spatial}
Sosulski, J. and Tangermann, M.
\newblock Spatial filters for auditory evoked potentials transfer between
  different experimental conditions.
\newblock In \emph{Proceedings of the 8th {Graz} Brain-Computer Interface
  Conference 2019}, pp.\  273--278, 2019.

\bibitem[Sosulski et~al.(2021)Sosulski, Kemmer, and
  Tangermann]{sosulski2021improving}
Sosulski, J., Kemmer, J.-P., and Tangermann, M.
\newblock Improving covariance matrices derived from tiny training datasets for
  the classification of event-related potentials with linear discriminant
  analysis.
\newblock \emph{Neuroinformatics}, 19\penalty0 (3):\penalty0 461--476, 2021.

\bibitem[Xiao et~al.(2012)Xiao, Wu, et~al.]{xiao2012covariance}
Xiao, H., Wu, W.~B., et~al.
\newblock Covariance matrix estimation for stationary time series.
\newblock \emph{Annals of Statistics}, 40\penalty0 (1):\penalty0 466--493,
  2012.

\bibitem[Zhao et~al.(2017)Zhao, Guo, Feng, and Chen]{zhao2017oracle}
Zhao, M., Guo, C., Feng, C., and Chen, S.
\newblock Oracle approximating shrinkage estimator based cooperative spectrum
  sensing for dense cognitive small cell network.
\newblock In \emph{2017 IEEE/CIC International Conference on Communications in
  China (ICCC)}, pp.\  1--6. IEEE, 2017.

\end{thebibliography}
\bibliographystyle{preprint_bib}

\newpage
\appendix
\onecolumn

\section{Detailed ERP Dataset Overview}
\label{sec:supp:datasets}

All subjects participated in one session, unless the dataset description denotes otherwise.

\subsection{Public Datasets}

\textbf{V\_BNCI\_A}: Dataset of 8 healthy subjects recorded at 8 channels performing a visual spelling task. The original dataset contains 10 subjects, but the first two have a different target/non-target ratio (1:35) as described in the full-text (1:5) and therefore were omitted from the benchmark.~\cite{guger2009how}

\textbf{V\_BNCI\_B}: Dataset of 8 ALS patients recorded at 8 channels performing a visual spelling task.~\cite{arico2014influence}

\textbf{V\_BNCI\_C}: Dataset of 10 healthy subjects recorded at 16 channels performing a visual spelling task. Note that each subject attended 4 recording sessions.~\cite{arico2014influence}

\textbf{V\_EPFL}: Dataset of 3 wheelchair-bound and 4 healthy subjects recorded at 32 channels performing a visual six-choice image selection task. Note that one wheelchair-bound subject had an excessive number of artifacts which lead to numerical errors in our pipeline and was omitted.~\cite{hoffmann2008efficient}

\textbf{V\_Lee}: Dataset of 54 healthy subjects recorded at 62 channels performing, among others, a visual spelling task.~\cite{lee2019eeg}

\textbf{V\_LLP\_A}: Dataset of 13 healthy subjects recorded at 31 channels performing a visual spelling task with a modified BCI paradigm that allows learning from label proportions. Each subject took part in three blocks and spelled 63 letters per block.~\cite{hubner2017learning}

\textbf{V\_LLP\_B}: Dataset of 12 healthy subjects recorded at 31 channels performing a visual spelling task with a modified BCI paradigm that allows learning from label proportions. Each subject took part in three blocks and spelled 70 letters per block, with the first 35 letters being instructed and the last 35 free spelling.~\cite{hubner2018unsupervised}

\textbf{V\_BI}: Dataset of 24 healthy subjects recorded at 16 channels performing a visual selection task based on the BrainInvaders game. Note that the first 7 subjects took part in 8 sessions each.~\cite{barachant2014plugplay}

\textbf{T\_Odd\_Young}: Dataset of 13 healthy subjects recorded at 31 channels performing an auditory tone oddball task recorded with different stimulation speeds.~\cite{sosulski2019spatial}

\subsection{Private Datasets}

\textbf{T\_Odd\_Healthy}: Dataset of 20 healthy elderly subjects recorded at 63 channels performing an auditory tone oddball task.~\cite{denzer2016influence}

\textbf{T\_Odd\_Stroke}: Dataset of 14 stroke patients recorded at 31 channels performing an auditory tone oddball task. Subjects performed multiple sessions.~\cite{musso2022aphasia}

\textbf{W\_Healthy}: Dataset of 20 healthy elderly subjects recorded at 63 channels performing an auditory oddball with word stimuli.~\cite{denzer2016influence}

\textbf{W\_Stroke}: Dataset of 10 stroke patients recorded at 31 channels performing an auditory oddball task with word stimuli. Subjects performed multiple sessions and we report averaged AUC performance.~\cite{musso2022aphasia}

\section{Evaluated Hyperparameters}
\label{sec:supp:hyperpars}

For LDA classification, ERP signals are often averaged in certain time intervals.
While these are often very specific to a recorded dataset, we want to avoid overfitting and evaluate on all datasets the following averaging interval boundaries, i.e., boundaries of $\{100, 200, 300\}$\,ms yield the averaging intervals $[100, 200)$\,ms and $[200, 300)$\,ms.

\begin{enumerate}
	\item $\{100, 170, 230, 300, 410, 500\}$\,ms ($N_t=5$)
	\item $\{100, 140, 170, 200, 230, 270, 300, 350, 410, 450, 500\}$\,ms ($N_t=10$)
	\item $\{0, 50, 100, 150, 200, 250, 300, 350, 400, 450, 500, 550, 600, 650, 700, 750, 800, 850, 900, 950, 1000\}$\,ms ($N_t=20$)
\end{enumerate}

As the optimal time intervals are often very paradigm specific, we evaluate two additional feature settings, where we simply use every sample (at 40\,Hz sampling rate) in the following two intervals.

\begin{enumerate}
	\item $[100, 600)$\,ms ($N_t = 20$)
	\item $[0, 1000)$\,ms ($N_t = 40$)
\end{enumerate}

While using all samples may generalize better across datasets, the high dimensionality poses an extra challenge for the classifiers.

The Riemannian geometry based classifier we used required an xDAWN decomposition~\cite{rivet2009xdawn} to enhance the ERP response and to reduce dimensionality.
We evaluated each of $\{1, 3, 6, 9\}$ as the number of used xDAWN components.
The Riemannian classifier can be trained using additional templates of either target ERPs only, or of target and non-target ERPs.
As performance depends on both of these parameters, we trained a classifier for each of the resulting eight possible combinations.

\section{ToeplitzLDA Performance Difference for each Dataset}
\label{sec:supp:benchmark_results}

\begin{figure}
	\centering
	\includegraphics[width=\textwidth]{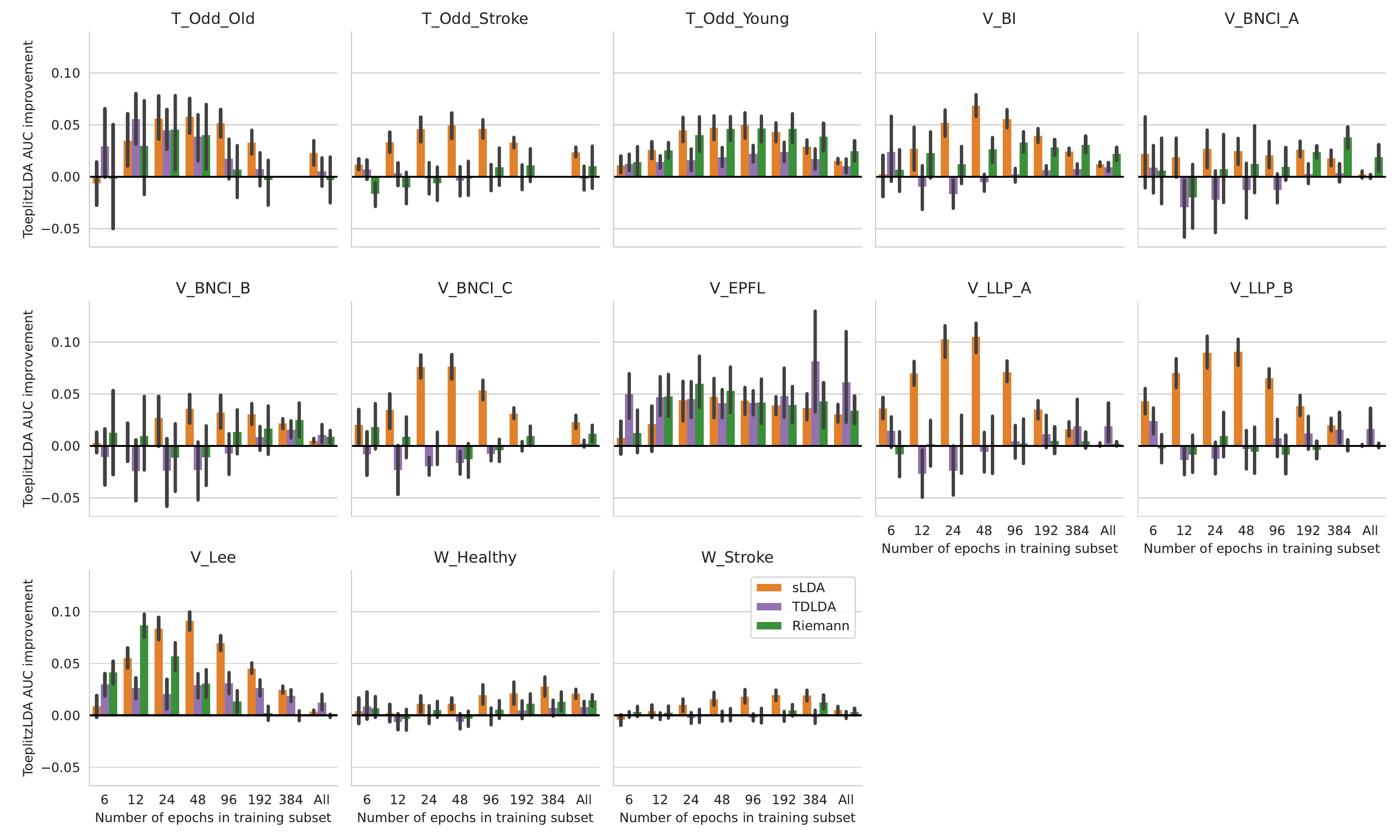}
	\caption{Improvement of classification performance in AUC of ToeplitzLDA compared to all other classifiers on all evaluated ERP datasets. The results are averaged over all subjects of a dataset. Error bars indicate the 95\,\% confidence interval. All classifiers were trained on the training subset only, i.e., for the LDA methods both class means and covariance were estimated on the subset.}
	\label{fig:supp:benchmark_results}
\end{figure}

\Cref{fig:supp:benchmark_results} shows the classification performance improvement of using ToeplitzLDA for each dataset.
Most notably in the word datasets, all classifiers performed on a similar, though low level. This can be explained by the general difficulty of the task, and as auditory evoked responses tend to be lower than visually evoked responses.

\section{Subject-wise Results for the Unsupervised P300 Speller}
\label{sec:supp:appl_results}

\begin{figure}
	\begin{subfigure}[c]{0.48\textwidth}
		\includegraphics[width=\textwidth]{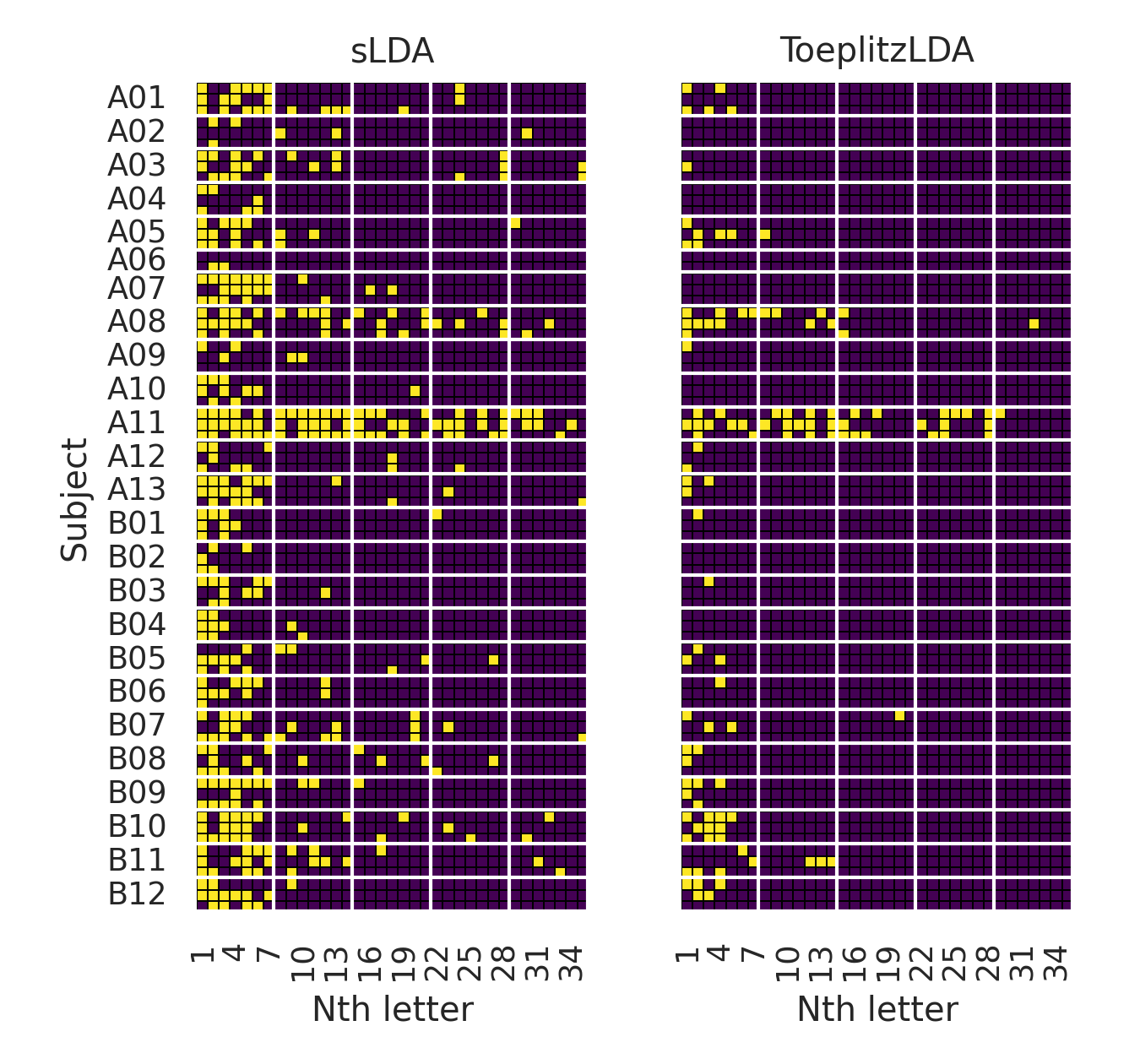}
		\subcaption{Heatmap showing classification results for the first 35 letters for each subject(left: sLDA, right: ToeplitzLDA). Dark purple squares indicate correctly selected letters and yellow blocks indicate wrong letters.}
		\label{fig:supp:heatmap}
	\end{subfigure}
	\hspace{0.04\textwidth}
	\begin{subfigure}[c]{0.48\textwidth}
		\includegraphics[width=\textwidth]{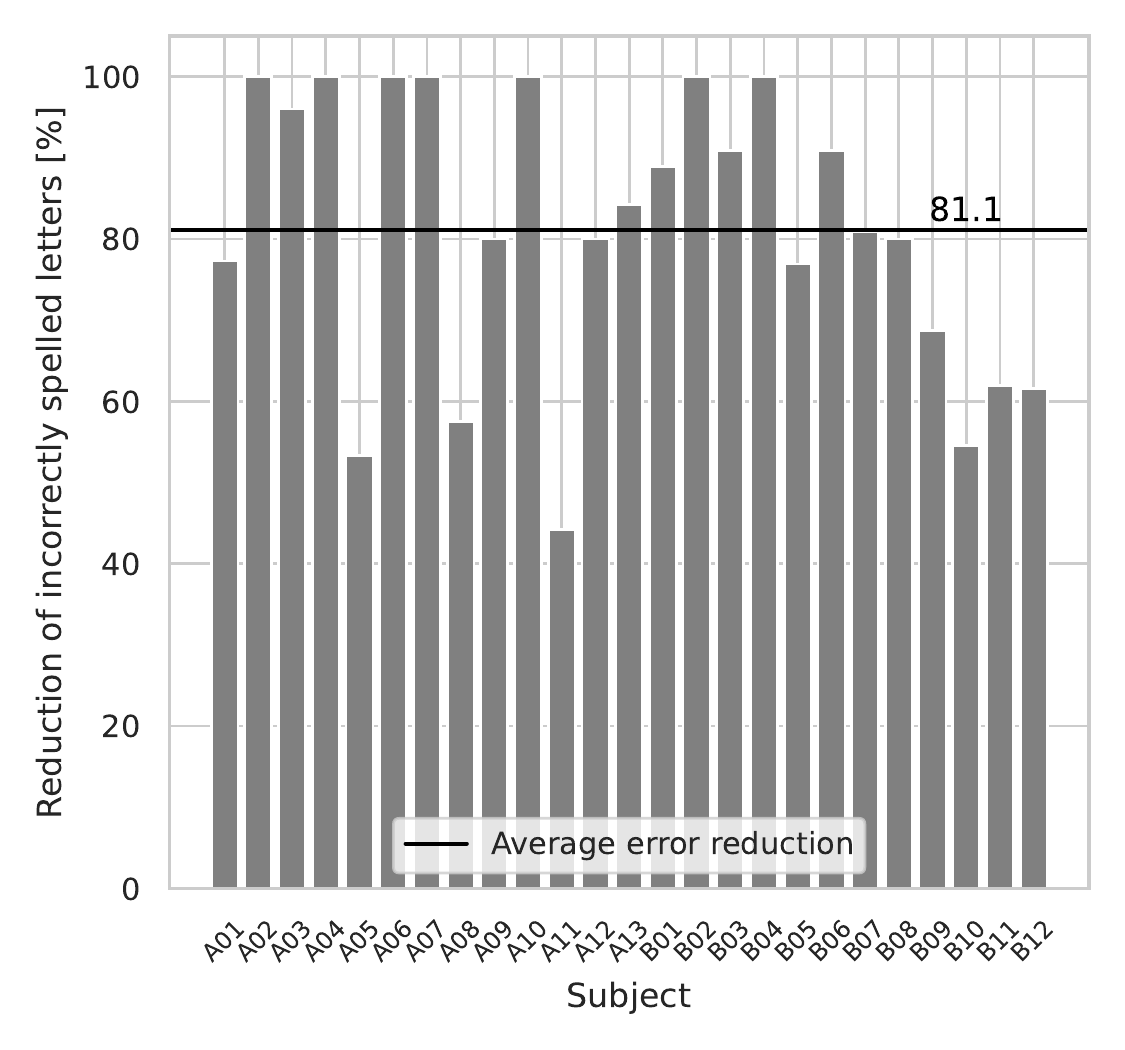}
		\subcaption{Reduction of incorrectly spelled letters for each subject for ToeplitzLDA compared to sLDA. The average reduction is indicated by the black line. For seven subjects, the ToeplitzLDA allows for a correct selection from the first letter on.}
		\label{fig:supp:barplot}
	\end{subfigure}
	\caption{Detailed results of the unsupervised visual speller application on both LLP datasets. Subjects from V\_LLP\_A are denoted with a leading `A' and analogously for V\_LLP\_B.}
	\label{fig:supp:usup}
\end{figure}

For fair performance comparisons, we used the best performing hyperparameters of sLDA ($N_t=6$) and of ToeplitzLDA ($N_t=27$) for the visualization of spelling performances in~\Cref{fig:supp:heatmap}. 
Note that we omitted the second block of subject 6 of dataset V\_LLP\_A, as the data could not be loaded using the provided dataset wrapper.

As shown in~\Cref{fig:supp:heatmap}, most errors of both methods occur during early letters and that the same subset of subjects can be considered difficult for both methods.
However, ToeplitzLDA manages to reduce many of the early errors and for most subjects becomes very stable once the classifier has seen enough data.
On average, ToeplitzLDA allows to reduce the number of incorrectly spelled letters by 81.1\,\% as shown in~\Cref{fig:supp:barplot}.
Notably, no single subject decreases in performance when using ToeplitzLDA instead of sLDA.

\end{document}